\pdfoutput=1

\documentclass[11pt]{article}
\usepackage{xcolor}
\usepackage{array}
\usepackage{multirow}
\usepackage{multicol}
\usepackage{graphicx}
\usepackage{float}
\usepackage{caption}
\usepackage{tabularx}
\usepackage{booktabs}
\usepackage{subcaption}
\usepackage{amsmath}
\usepackage{svg}

\usepackage[preprint]{acl}
\usepackage{enumitem}
\usepackage{times}
\usepackage{latexsym}

\usepackage[T1]{fontenc}

\usepackage[utf8]{inputenc}

\usepackage{microtype}

\usepackage{inconsolata}

\usepackage{graphicx}
\usepackage{longtable}

\newcommand{\bloomsae}{Bloom SAE}
\newcommand{\opensae}{Open AI SAE}
\newcommand{\apollosae}{Apollo SAE}
\newcommand{\gptsmall}{GPT-2 small}
\newcommand{\gptSmall}{GPT-2 Small}
\newcommand{\ReLU}{\mathsf{ReLU}}

\newcommand{\rotmat}{\mathbf{R}}

\newcommand{\model}{\mathcal{M}}
\newcommand{\base}{b}
\newcommand{\source}{s}
\newcommand{\y}{y}
\newcommand{\ypred}{\hat{y}}
\newcommand{\features}{\mathbf{F}}
\newcommand{\featurevals}{\mathbf{f}}
\newcommand{\getfeatures}[3]{\mathsf{get}(#1,#2, #3)}
\newcommand{\setfeatures}[3]{#1_{#2 \leftarrow #3}}

\newcommand{\h}{\mathbf{h}}

\title{Evaluating Open-Source Sparse Autoencoders \\on Disentangling Factual Knowledge in \gptSmall}

\author{Maheep Chaudhary \\
  Pr(Ai)$^{2}$R Group\\
\texttt{maheepchaudhary.research@gmail.com} \\ \AnD
  Atticus Geiger\\
  Pr(Ai)$^{2}$R Group\\
  \texttt{atticusg@gmail.com}\\}

\begin{document}
\maketitle
\begin{abstract}
A popular new method in mechanistic interpretability is to train high-dimensional sparse autoencoders (SAEs) on neuron activations and use SAE features as the atomic units of analysis. However, the body of evidence on whether SAE feature spaces are useful for causal analysis is underdeveloped. In this work, we use the RAVEL benchmark to evaluate whether SAEs trained on hidden representations of \gptsmall\ have sets of features that separately mediate knowledge of which country a city is in and which continent it is in. We evaluate four open-source SAEs for \gptsmall\ against each other, with neurons serving as a baseline, and linear features learned via distributed alignment search (DAS) serving as a skyline. For each, we learn a binary mask to select features that will be patched to change the country of a city without changing the continent, or vice versa. Our results show that SAEs struggle to reach the neuron baseline, and none come close to the DAS skyline. 
We release code here: 
\href{https://github.com/MaheepChaudhary/SAE-Ravel}{github.com/MaheepChaudhary/SAE-Ravel}
\end{abstract}

\section{Introduction}

Individual neurons in neural networks represent many concepts, and individual concepts are represented by many neurons \cite{polysemantic7, polysemantic6, polysemantic5, polysemantic4, polysemantic3, polysemantic2, polysemantic1}. What, if not neurons, are the relevant meaning-bearing components of neural networks? This is a fundamental question in mechanistic interpretability. A recent, and increasingly popular, unsupervised method for learning features that correspond to intuitive concepts is to train high-dimensional sparse autoencoders (SAEs) on the hidden representations of deep learning models across a wide range of possible inputs~\cite{sae1, cunningham2023sparseautoencodershighlyinterpretable, lieberum2024gemmascopeopensparse, Gao:2024}. The encoder of an SAE unpacks neurons into a higher dimensional space with sparse linear features that are intended to be better units of analysis.

However, researchers have invested more into scaling SAEs, than evaluating them \cite{templeton2024scaling}. In particular, only a handful of works engage with whether SAEs are useful for a causal interpretability analysis \cite{marks2024,Engels2024,makelov2024}. In this paper, we add to the body of evidence an example of when sparse autoencoders fail to provide a better feature space than neurons for finding model-internal mediators of concepts \cite{geiger2024causalabstractiontheoreticalfoundation, mueller2024questrightmediatorhistory}. Specifically, we use the RAVEL benchmark \cite{ravel} to evaluate whether the there are sets of SAE features that separately mediate knowledge of which country a city is in and which continent a city is in. We evaluate four publicly available SAEs for \gptsmall: the \opensae\ \cite{Gao:2024}, two \apollosae s \cite{e2eapollo}, and the \bloomsae\ \cite{indepsae}. As a feature baseline, we use neurons; as a feature skyline, we use linear subspaces trained with distributed alignment search (DAS; \citealt{DAS}) to disentangle the country knowledge from the continent knowledge.

\text{For each feature space, we train a differentiable} binary mask to select features that encode the country of a city, but not the continent, and vice versa. We evaluate the selected features using interchange interventions, where features are fixed to values they would take if a different input were provided. For example, if we fix the `country' features for the prompt \textit{Paris is a city in the country} and set them to the value they take for the prompt \textit{Tokyo is a city in}, the output should be \textit{Japan} not \textit{France}. If we instead target the `continent' features, the output should be \textit{Europe} not \textit{Asia}.

\begin{figure*}[hbtp]
    \centering
\begin{subfigure}{\textwidth}
    \centering
    \includegraphics[width=0.495\textwidth]{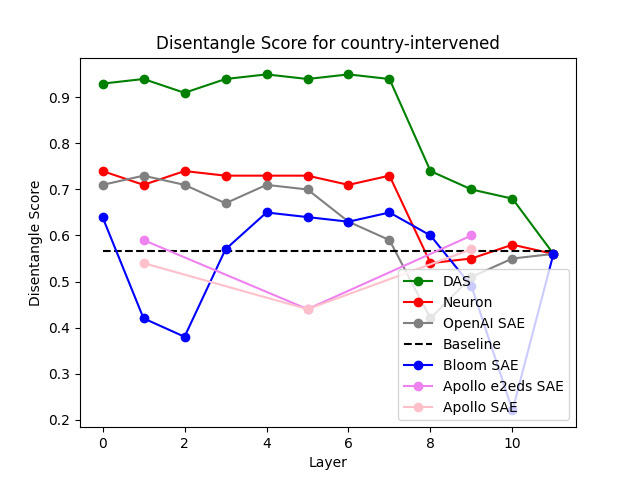}
    \includegraphics[width=0.495\textwidth]{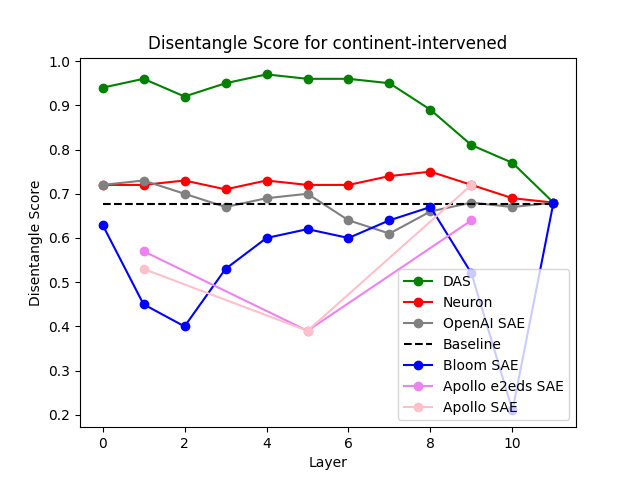}
    \caption{The disentangle score for `continent' and `country' interventions across the layers of \gptsmall. The disentangle scores for empty intervention baselines are shown as dotted lines. The performance of the DAS skyline goes down after layer 7 because the knowledge about the city is being moved away from the \texttt{<city>} token. The \apollosae s are only available for layers 1, 5, and 9. }
    \label{fig:plot}
\end{subfigure}
        
\hspace{400pt}
\begin{subfigure}{\textwidth}
        \centering
        \resizebox{\textwidth}{!}{
    \begin{tabular}{c c c c c c c || c c c c c c c c}
    \toprule
    & \multicolumn{6}{c||}{\textbf{Country-Intervened Continent-Preserved}} & \multicolumn{6}{c}{\textbf{Continent-Intervened Country-Preserved}} \\
    & {Neurons} & {DAS} & Open AI & Bloom & Apollo & Apollo & Neuron & DAS & Open AI & Bloom & Apollo & Apollo \\
    & & & SAE & SAE & SAE e2e & SAE e2e+ds & & & SAE & SAE & SAE e2e & SAE e2e+ds \\
    \midrule
    Continent Accuracy & $46$ & $93$ & $51$ & $36$ & $24$ & $33$ & $48$ & $94$ & $49$ & $37$ & $24$ & $32$ \\
    {Country Accuracy} & $96$ & $94$ & $95$ & $49$ & $84$ & $86$ & $97$ & $99$ & $97$ & $52$ & $81$ & $82$ \\
    {Disentangle Score} & $71$ & $\textbf{94}$ & $73$ & $43$ & $54$ & $59$ & $72$ & $\textbf{96}$ & $73$ & $45$ & $52$ & $57$ \\
    \midrule
    {Inactive Features} & $0$ & $0$ & $0.977$ & $0.98$ & $0.966$ & $0.974$ & $0$ & $0$ & $0.977$ & $0.98$ & $0.966$ & $0.974$ \\
    {Non-Intervened Features} & $0.11$ & $0.24$ & $0.006$ & $0.005$ & $0.01$ & $0.009$ & $0.88$ & $0.79$ & $0.015$ & $0.009$ & $0.023$ & $0.018$ \\
    {Intervened Features} & $0.89$ & $0.76$ & $0.015$ & $0.009$ & $0.023$ & $0.018$ & $0.12$ & $0.21$ & $0.007$ & $0.005$ & $0.011$ & $0.009$ \\
    \midrule
        Reconstruction Loss & 0 & 0 & 152 & 551 & 2245 & 2130 & 0 & 0 & 158 & 516 & 2576 & 2318 \\ 
    Reconstructed Knowledge & 100 & 100 & 95 & 56 & 67 & 1 & 100 & 100 & 95 & 47 & 35 & 0 \\
    \bottomrule
    \end{tabular}
    } 
    \caption{GPT-2 small at layer 1. The first three rows are interchange intervention accuracies for RAVEL using learned binary masks to select features. The next three rows are sparsity evaluations that show the proportion of inactive features, intervened on features, and active non-intervened features. The final two rows are reconstruction evaluations that show the models knowledge of cities using a reconstructed representation (no interventions performed) and the average mean-squared error reconstruction loss. The base prompts for each of the two datasets were used for reconstruction evaluations, with source prompts being ignored.}
    \label{tab:Layer1}
        \end{subfigure}
        
\hspace{400pt}
    \begin{subfigure}{\textwidth}
    \resizebox{\textwidth}{!}{
    \begin{tabular}{c c c c c c c || c c c c c c}
        \toprule
         & \multicolumn{6}{c||}{\textbf{Country-Intervened Continent-Preserved}} & \multicolumn{6}{c}{\textbf{Continent-Intervened Country-Preserved}} \\
        & {Neurons} & {DAS} & Open AI & Bloom & Apollo & Apollo & Neuron & DAS & Open AI & Bloom & Apollo & Apollo\\
        &  & & SAE & SAE & e2e+ds SAE & SAE & & & SAE & SAE & e2e+ds SAE & SAE\\
        \midrule
        Continent Accuracy & $49$ & $91$ & $53$ & $48$ & $22$ & $22$ & $46$ & $93$ & $48$ & $45$ & $18$ & $18$ \\
        {Country Accuracy} & $97$ & $98$ & $88$ & $79$ & $66$ & $66$ & $98$ & $99$ & $91$ & $79$ & $61$ & $61$ \\
        {Disentangle Score} & $73$ & $\textbf{94}$ & $70$ & $64$ & $44$ & $44$ & $72$ & $\textbf{96}$ & $70$ & $62$ & $40$ & $39$ \\
        \midrule
        {Inactive Features}  & $0$ & $0$ & $0.951$ & $0.979$ & $0.981$ & $0.98$ & $0$ & $0$ & $0.951$ & $0.979$ & $0.981$ & $0.98$ \\
        {Non-Intervened Features}& $0.119$ & $0.325$ & $0.017$ & $0.006$ & $0.008$ & $0.009$ & $0.877$ & $0.69$ & $0.031$ & $0.013$ & $0.017$ & $0.017$ \\
        {Intervened Features} & $0.88$ & $0.674$ & $0.031$ & $0.013$ & $0.017$ & $0.017$ & $0.122$ & $0.309$ & $0.017$ & $0.006$ & $0.008$ & $0.009$\\
        \midrule
        Reconstruction Loss & 0 & 0 & 644 & 937 & 2383 & 2353 & 0 & 0 & 652 & 1044 & 2576 & 2318\\  
        Reconstructed Knowledge & 100 & 100 & 88 & 77&84 & 83  & 100 & 100 & 90 & 86 & 76 & 61 \\
        \bottomrule
    \end{tabular}
    } 
    \caption{GPT-2 small at layer 5. See the caption above from Figure~\ref{tab:Layer1} for details on the table structure.}
    \label{tab:Layer5}
        \end{subfigure}
    \caption{Metrics on the RAVEL test set for interchange interventions performed on the residual stream in \gptsmall\ after transformer block above the city token \texttt{<city>}. For each space of features, we learn `country' features that encode what country a city is in and `continent' features that encode what continent a city is in. Interventions targeting the `country' features should change the output for the prompt \texttt{<city>}\textit{ is in the country of}, but not \texttt{<city>}\textit{ is in the continent of}. Interventions targeting the `continent' features should do the opposite. The disentangle score is the average of the country and continent accuracies. Neurons serve as a baseline for how easily these two facts are disentangled, and DAS is a supervised feature learning method that serves as a skyline. The SAEs are the methods we seek to evaluate. In sum, using SAE reconstructions harm the knowledge of GPT-2 and SAE features are not better mediators than the baseline of neurons.}
    \label{fig:mainresults}
\end{figure*}

In Figure~\ref{fig:mainresults} we show that all SAEs struggle to compete with the neuron baseline and degrade the model's knowledge. However, the DAS skyline sets a high ceiling and there is room to improve.

\section{Related Work}
\paragraph{Benchmarking SAEs}
There are many aspects of SAEs to benchmark. To what degree do the features respond precisely and accurately to the natural language labels given to them by auto-interpretability methods \cite{Hernandez:2022NaturalLanguage,Huang:2023,Schwettmann2023FIND,bills2023language, Shaham2024}? Can we do circuit discovery \cite{marks2024, makelov2024}, representation analysis \cite{Engels2024}, or activation steering \cite{templeton2024scaling} in SAE feature space? Our question is whether SAEs provide a better feature space than neurons for localizing the concepts used by deep learning models.

\paragraph{Interpretability of Knowledge Representations}
The RAVEL benchmark belongs to a line of research concerned with how factual knowledge is stored within a language model \cite{geva-2021-transformer, Meng:2022, dai-2022-knowledge, Meng:2023,  hernandez2023measuring,geva2023dissecting}. In this paper, we are concerned with how factual knowledge is stored and processed in hidden vector representations during model inference. Activation steering or model editing ask how to control a model, whereas we ask how a model constructs and manipulates representations to control itself.

\section{Methodology}

\subsection{The RAVEL Benchmark}\label{sec:ravel}
RAVEL \cite{ravel} is an benchmark that evaluates interpretability methods on localizing and disentangling related factual knowledge. We focus on the data for disentangling the country a city is in from the continent it is in. 

\paragraph{Filtering} Following \citet{ravel}, we filter out all of the cities that \gptsmall\ \cite{radford2019language} doesn't know both the country and the continent. However, \gptsmall\ is not a very capable model, so we give five in context examples when evaluating the knowledge of the model:

\emph{``Toronto is a city in the country of Canada. Beijing is a ... \texttt{<city>} is a city in the country of''}

See Appendix~\ref{app:eval} for the full 5-shot prompts. We further filter out multi-token cities to simplify the task and give the SAEs the best chance at success. The resulting dataset contains $40$ cities in total.

\paragraph{Interchange Interventions in Feature Space}
If a set of features contains the knowledge that \textit{Toronto} is in \textit{Canada}, then fixing those features to take on the value they would have for city \textit{Tokyo} should make the model think that \textit{Toronto} is in \textit{Japan}. The process of fixing features to take on values they would have for a different input is an \textit{interchange intervention} \cite{geiger-etal:2020:blackbox, vig2020, finlayson-etal-2021-causal}. 
Suppose we have base input prompt $\base$ and a source input prompt $\source$ for a model $\model$ and we want to target features $\features$. Define the interchange intervention as
\begin{align*}
\featurevals = \getfeatures{\model}{\source}{\features} \\
\ypred = \setfeatures{\model}{\features}{\featurevals}(\base) 
\end{align*}
where $\getfeatures{\model}{\source}{\features}$ retrieves the value that features $\features$ take on when $\model$ is run with input $\source$ and $\setfeatures{\model}{\features}{\featurevals}(\base)$ is the output produced when $\model$ is run with input $\base$ under intervention $\features \leftarrow \featurevals$. 

\paragraph{Counterfactual Labels} The label of an interchange intervention example is determined by the concept we think is encoded in the features $\features$ and the mechanism that determines the output given the prompt \cite{geiger2021}. For our task, the mechanism connecting the knowledge of a city and the expected behavior is simple. If we are intervening on the `country' features, then the `country' prompt should have the label from the source $\y_{\source}$ and the `continent' prompt should have the label from the base $\y_{\base}$. If we intervene on the `continent' feature, we use the opposite labels.

\paragraph{Splits} To evaluate a proposed set of `country' features and `continent' features, we perform interchange interventions using the RAVEL dataset prompts for base and source inputs. We filtered our dataset down to $40$ cities, which can be used to generate $1600$ interchange interventions targeting `country' and $1600$ interchange interventions targeting `continent' ($3200$ in total). We split the interchange intervention data so that 70\% is training, 10\% is validation, and 20\% is test. Our evaluations are i.i.d. to give SAEs the best chance at success.

\subsection{Constructing and Selecting Features }
\paragraph{Sparse Autoencoders for Dictionary Learning}
Sparse autoencoders (SAEs;\citealt{sae1, cunningham2023sparseautoencodershighlyinterpretable}) are a unsupervised method for unpacking a hidden vector representation into a higher dimensional, sparsely activated feature space. The hope is that dimensions in this new feature space will correspond to interpretable concepts. SAEs used for this purpose typically have an encoder with a linear transformation followed by a $\ReLU$\ and a linear decoder:
\begin{align*}
\bar{x} &= x - b_x \\
f &= \ReLU(W_e \bar{x} + b_e) \\
\hat{x} &= W_d f + b_d 
\end{align*}

SAEs are optimized jointly to have low reconstruction error and sparse representations:
\[\mathcal{L} = \frac{1}{|X|} \sum_{x \in X} \left\| x - \hat{x} \right\|_2^2 + \lambda \left\| f \right\|_1\]
Low reconstruction loss ensures that the features faithful to the underlying hidden vector and low sparsity loss is thought to create interpretable features. General purpose SAEs are trained on hidden vector representations created by the model when processing a enormous amount of text data, e.g., an SAE might be trained on residual stream representations created by the second layer of a transformer processing the Pile \cite{Gao2021Pile}.

The \bloomsae\ has this standard architecture and training, but the other SAEs are variants. The \opensae\ is a top-k SAE, which \cite{Gao:2024} show to outperform the standard architecture on the sparsity-reconstruction frontier.  A top-$k$ encoder is simply the standard encoder except only the top-$k$ firing features are kept:
\[f = \mathsf{Topk}(\ReLU(W_e \bar{x} + b_e)) \]

The two \apollosae s have standard architecture, but they are trained with additional loss terms. The \apollosae\ (e2e) is trained with the additional loss objective of the KL-divergence between the output logits of the model before and after reconstruction. The \apollosae\ (e2e + ds) has the logit-based loss in addition to a mean-squared error loss between the residual stream representations in downstream layers before and after reconstruction. \cite{e2eapollo} also report a praeto improvement on the sparsity-reconstruction trade off for end-to-end models.

\paragraph{Distributed Alignment Search}
SAEs are unsupervised, so features must be further analyzed to determine their conceptual content. In contrast, DAS \cite{DAS} learns linear features with specific conceptual content via supervision from counterfactual data that describes how a model should act when a concept has been intervened upon. DAS features learned specifically for this task will be a skyline for general-purpose SAEs.

In particular, DAS learns an orthogonal matrix $\rotmat$ that rotates a hidden vector $\h$, with the dimensions of the rotated space $\rotmat \h$ being the new feature space, i.e. a set of features $\features$ are dimensions of $\rotmat \h$. We start by randomly initializing $\rotmat$, which renders all features equally meaningless. Then, an interchange intervention is performed on features $\features$ with a base $\base$ and source $\source$ input prompt pair. Loss is computed from the output of the intervened model:

\begin{align*}
\mathcal{L} &= \text{CE}( \setfeatures{\model}{\features}{\getfeatures{\model}{\source}{\features}}(\base), \y)
\end{align*}
The expected label $\y$ is determined by the concept that we are localizing in $\features$ and the mechanism by which the concept determines behavior. See Section~\ref{sec:ravel} for a description of the interchange intervention data. We provide details on hyperparameters in Appendix~\ref{app:hyper}.

\paragraph{Differential Binary Masking}
In order to determine which features $\features$ to select for a given concept (`country' and `continent' in our case), we use Differential Binary Masking (DBM;~\citealt{DBM4, DBM3, DBM2, DBM1}) to select features for intervention. Each feature $f$ in the feature space $\mathcal{F}$ is masked with a vector $\mathbf{m}$ which is passed into a sigmoid $\sigma$ after being scaled by a temperature $T$:
\begin{align*}
\featurevals_{\base} &= \getfeatures{\model}{\features}{\base}\\
\featurevals_{\source}& = \getfeatures{\model}{\features}{\source}\\
\featurevals &= (1 - \sigma(\mathbf{m}/T)) \odot \featurevals_{\base} + \sigma(\mathbf{m}/T) \odot \featurevals_{\source} 
\end{align*}
These masks are trained on an interchange intervention loss objective while the temperature is annealed to make the masks snap to 0 or 1:
\[\mathcal{L} = \text{CE}( \setfeatures{\model}{\features}{\featurevals}(\base), \y)\]
When we DBM with DAS, the features and the masks are learned simultaneosly.

\section{Experiments}
Our goal is to find a hidden vector representation in \gptsmall\ where the DAS skyline features are significantly better than the neuron baseline, and then evaluate whether SAEs are an improvement on neurons as a unit of analysis. For this reason, we follow the lead of \cite{Huang:2023} and chose to explore the residual stream representations of \gptsmall\ above the \texttt{<city>} token in the early layers of the model.
We implement our experiments with nnsight \cite{fiottokaufman2024nnsightndifdemocratizingaccess} and pytorch \cite{pytorch}.

\subsection{Results}
In Figure~\ref{fig:plot}, we report the interchange intervention accuracy across the layers of \gptsmall.  In Figures~\ref{tab:Layer1} and~\ref{tab:Layer5}, we present the detailed results for layers 1 and 5 of \gptsmall , because the \apollosae s are available for those two layers. We learned `country' features and `continent' features, then we used interchange interventions on those features to evaluate whether they, in fact, store the model's knowledge of the country and continent that a city is in, respectively. When targeting `country' features for intervention, the `country' accuracy is high when the intervention changes the output and the `continent' accuracy is high when the intervention does not change the model output. The opposite is true for interventions on `continent' features. The `disentangle score' is the average of the two accuracies. In the middle three rows of the table are sparsity evaluations that report how many features were active and/or intervened upon. In the final two rows of the table are reconstruction evaluations that report the knowledge degradation of \gptsmall\ when a reconstructed vector is used and the average reconstruction loss on residual stream vectors above the \texttt{<city>} token at a given layer.

\subsection{Discussion}

\paragraph{Using representations reconstructed by SAEs degrades the model's knowledge of cities.} The last row in Figures~\ref{tab:Layer1} and~\ref{tab:Layer5} that using a representation reconstructed by an SAE always degrades the model's knowledge of the countries and continents that cities belong to. For the first layer, we can see that the \bloomsae\ and \apollosae e2e severely harm the model ($\approx$ -50\%) and the \apollosae\ e2e+ds destroys the knowledge entirely. In contrast, the \opensae\ results in only a small drop in performance (-5\%). For the fifth layer, there is less degradation, the \apollosae\ e2e+ds works,  and \opensae\ is again the best.

\paragraph{The end-to-end SAEs degrade knowledge less relative to the reconstruction loss.} In our limited evaluations, there is no evidence that end-to-end training used to create the two \apollosae s was helpful for providing a feature space where knowledge can be disentangled. However, in the last two rows of Figure~\ref{tab:Layer5} we can see that despite have the highest reconstruction loss, the \apollosae (e2e + ds) degrades the city-knowledge of \gptsmall\ an amount that is comparable with \opensae\ and \bloomsae . This is weak evidence that the end-to-end objective was helpful for preserving model capabilities.

\paragraph{There is a signifigant gap between baseline and skyline; neurons can be improved upon.} The skyline provided by DAS at $\approx$95\% accuracy for the first 7 layers of \gptsmall\ shows that there are separate linear subspaces that encode the country a city is in and the continent a city is in. This means, an SAE with linear features that span these subspaces could achieve performance equivalent to DAS. The neuron baseline at $\approx$70\% is significantly worse than the DAS skyline, and shows that there are polysemantic neurons that need to be disentangled by a rotation via an orthogonal matrix. 

\paragraph{Current SAEs for \gptsmall\ struggle to compete with the neurons.} The two \apollosae s and \bloomsae\ below the neuron baseline across all layers. The `country' and `continent' knowledge are even more entangled in the feature spaces provided by these SAEs. The \opensae\ at $\approx$70\% is able to match the performance of the neuron baseline, but not exceed it. 

\paragraph{The top-k SAE is the most performant.} Our evaluation is limited, however the results do seem to track improvements in SAEs.  The \opensae\ is a top $k$-SAE, which a performant architecture on sparsity and reconstruction evaluations \cite{Gao:2024}. This is in line with our results that the \opensae\ is the only model that competes with the neuron baseline across all layers.

\section{Conclusion}

We evaluate open-source SAEs on their ability to provide a feature space for GPT-2 hidden representations where knowledge about the country and continent a city is in can be disentangled. We used neurons as a baseline feature space, and a supervised feature learned by DAS as a skyline feature space. While we were able to see meaningful differences in performance between the three SAEs, only one of the evaluated SAEs was able to reach the neuron baseline and none could reach the DAS skyline. We hope this is a useful step in evaluating the usefulness of SAEs for a causal interpretability analysis of deep learning models.

\section*{Limitations}

In future, we would like to scale the experiments to models with available SAEs including gemma, Mistral, Llama, and Pythia. 
Furthermore, we hope to use more attributes from the RAVEL dataset, such as \emph{language, gender}, etc. for larger models with more knowledge.

\bibliography{custom}

\begin{thebibliography}{44}
\providecommand{\natexlab}[1]{#1}

\bibitem[{Bills et~al.(2023)Bills, Cammarata, Mossing, Tillman, Gao, Goh, Sutskever, Leike, Wu, and Saunders}]{bills2023language}
Steven Bills, Nick Cammarata, Dan Mossing, Henk Tillman, Leo Gao, Gabriel Goh, Ilya Sutskever, Jan Leike, Jeff Wu, and William Saunders. 2023.
\newblock Language models can explain neurons in language models.
\newblock \url{https://openaipublic.blob.core.windows.net/neuron-explainer/paper/index.html}.

\bibitem[{Bloom(2024)}]{indepsae}
Joseph Bloom. 2024.
\newblock \href {https://www.alignmentforum.org/posts/f9EgfLSurAiqRJySD/open-source-sparse-autoencoders-for-all-residual-stream} {Open source sparse autoencoders for all residual stream layers of gpt2 small}.

\bibitem[{Bolukbasi et~al.(2021)Bolukbasi, Pearce, Yuan, Coenen, Reif, Viégas, and Wattenberg}]{polysemantic2}
Tolga Bolukbasi, Adam Pearce, Ann Yuan, Andy Coenen, Emily Reif, Fernanda Viégas, and Martin Wattenberg. 2021.
\newblock \href {https://arxiv.org/abs/2104.07143} {An interpretability illusion for bert}.
\newblock \emph{Preprint}, arXiv:2104.07143.

\bibitem[{Braun et~al.(2024)Braun, Taylor, Goldowsky-Dill, and Sharkey}]{e2eapollo}
Dan Braun, Jordan Taylor, Nicholas Goldowsky-Dill, and Lee Sharkey. 2024.
\newblock \href {https://arxiv.org/abs/2405.12241} {Identifying functionally important features with end-to-end sparse dictionary learning}.
\newblock \emph{Preprint}, arXiv:2405.12241.

\bibitem[{Bricken et~al.(2023)Bricken, Templeton, Batson, Chen, Jermyn, Conerly, Turner, Anil, Denison, Askell, Lasenby, Wu, Kravec, Schiefer, Maxwell, Joseph, Hatfield-Dodds, Tamkin, Nguyen, McLean, Burke, Hume, Carter, Henighan, and Olah}]{sae1}
Trenton Bricken, Adly Templeton, Joshua Batson, Brian Chen, Adam Jermyn, Tom Conerly, Nick Turner, Cem Anil, Carson Denison, Amanda Askell, Robert Lasenby, Yifan Wu, Shauna Kravec, Nicholas Schiefer, Tim Maxwell, Nicholas Joseph, Zac Hatfield-Dodds, Alex Tamkin, Karina Nguyen, Brayden McLean, Josiah~E Burke, Tristan Hume, Shan Carter, Tom Henighan, and Christopher Olah. 2023.
\newblock \href {https://transformer-circuits.pub/2023/monosemantic-features/index.html} {Towards monosemanticity: Decomposing language models with dictionary learning}.
\newblock \emph{Transformer Circuits Thread}.

\bibitem[{Cammarata et~al.(2020)Cammarata, Carter, Goh, Olah, Petrov, Schubert, Voss, Egan, and Lim}]{polysemantic3}
Nick Cammarata, Shan Carter, Gabriel Goh, Chris Olah, Michael Petrov, Ludwig Schubert, Chelsea Voss, Ben Egan, and Swee~Kiat Lim. 2020.
\newblock \href {https://doi.org/10.23915/distill.00024} {Thread: Circuits}.
\newblock \emph{Distill}.
\newblock Https://distill.pub/2020/circuits.

\bibitem[{Csord{\'a}s et~al.(2021)Csord{\'a}s, van Steenkiste, and Schmidhuber}]{DBM3}
R{\'o}bert Csord{\'a}s, Sjoerd van Steenkiste, and J{\"u}rgen Schmidhuber. 2021.
\newblock \href {https://openreview.net/forum?id=7uVcpu-gMD} {Are neural nets modular? inspecting functional modularity through differentiable weight masks}.
\newblock In \emph{International Conference on Learning Representations}.

\bibitem[{Cunningham et~al.(2023)Cunningham, Ewart, Riggs, Huben, and Sharkey}]{cunningham2023sparseautoencodershighlyinterpretable}
Hoagy Cunningham, Aidan Ewart, Logan Riggs, Robert Huben, and Lee Sharkey. 2023.
\newblock \href {https://arxiv.org/abs/2309.08600} {Sparse autoencoders find highly interpretable features in language models}.
\newblock \emph{Preprint}, arXiv:2309.08600.

\bibitem[{Dai et~al.(2022)Dai, Dong, Hao, Sui, Chang, and Wei}]{dai-2022-knowledge}
Damai Dai, Li~Dong, Yaru Hao, Zhifang Sui, Baobao Chang, and Furu Wei. 2022.
\newblock \href {https://aclanthology.org/2022.acl-long.581} {Knowledge neurons in pretrained transformers}.
\newblock In \emph{acl}.

\bibitem[{Davies et~al.(2023)Davies, Nadeau, Prakash, Shaham, and Bau}]{DBM1}
Xander Davies, Max Nadeau, Nikhil Prakash, Tamar~Rott Shaham, and David Bau. 2023.
\newblock \href {https://arxiv.org/abs/2307.03637} {Discovering variable binding circuitry with desiderata}.
\newblock \emph{Preprint}, arXiv:2307.03637.

\bibitem[{De~Cao et~al.(2020)De~Cao, Schlichtkrull, Aziz, and Titov}]{DBM4}
Nicola De~Cao, Michael~Sejr Schlichtkrull, Wilker Aziz, and Ivan Titov. 2020.
\newblock \href {https://doi.org/10.18653/v1/2020.emnlp-main.262} {How do decisions emerge across layers in neural models? interpretation with differentiable masking}.
\newblock In \emph{Proceedings of the 2020 Conference on Empirical Methods in Natural Language Processing (EMNLP)}, pages 3243--3255, Online. Association for Computational Linguistics.

\bibitem[{De~Cao et~al.(2022)De~Cao, Schmid, Hupkes, and Titov}]{DBM2}
Nicola De~Cao, Leon Schmid, Dieuwke Hupkes, and Ivan Titov. 2022.
\newblock \href {https://doi.org/10.18653/v1/2022.blackboxnlp-1.2} {Sparse interventions in language models with differentiable masking}.
\newblock In \emph{Proceedings of the Fifth BlackboxNLP Workshop on Analyzing and Interpreting Neural Networks for NLP}, pages 16--27, Abu Dhabi, United Arab Emirates (Hybrid). Association for Computational Linguistics.

\bibitem[{Engels et~al.(2024)Engels, Liao, Michaud, Gurnee, and Tegmark}]{Engels2024}
Joshua Engels, Isaac Liao, Eric~J. Michaud, Wes Gurnee, and Max Tegmark. 2024.
\newblock \href {https://doi.org/10.48550/ARXIV.2405.14860} {Not all language model features are linear}.
\newblock \emph{CoRR}, abs/2405.14860.

\bibitem[{Finlayson et~al.(2021)Finlayson, Mueller, Gehrmann, Shieber, Linzen, and Belinkov}]{finlayson-etal-2021-causal}
Matthew Finlayson, Aaron Mueller, Sebastian Gehrmann, Stuart Shieber, Tal Linzen, and Yonatan Belinkov. 2021.
\newblock \href {https://doi.org/10.18653/v1/2021.acl-long.144} {Causal analysis of syntactic agreement mechanisms in neural language models}.
\newblock In \emph{Proceedings of the 59th Annual Meeting of the Association for Computational Linguistics and the 11th International Joint Conference on Natural Language Processing (Volume 1: Long Papers)}, pages 1828--1843, Online. Association for Computational Linguistics.

\bibitem[{Fiotto-Kaufman et~al.(2024)Fiotto-Kaufman, Loftus, Todd, Brinkmann, Juang, Pal, Rager, Mueller, Marks, Sharma, Lucchetti, Ripa, Belfki, Prakash, Multani, Brodley, Guha, Bell, Wallace, and Bau}]{fiottokaufman2024nnsightndifdemocratizingaccess}
Jaden Fiotto-Kaufman, Alexander~R Loftus, Eric Todd, Jannik Brinkmann, Caden Juang, Koyena Pal, Can Rager, Aaron Mueller, Samuel Marks, Arnab~Sen Sharma, Francesca Lucchetti, Michael Ripa, Adam Belfki, Nikhil Prakash, Sumeet Multani, Carla Brodley, Arjun Guha, Jonathan Bell, Byron Wallace, and David Bau. 2024.
\newblock \href {https://arxiv.org/abs/2407.14561} {Nnsight and ndif: Democratizing access to foundation model internals}.
\newblock \emph{Preprint}, arXiv:2407.14561.

\bibitem[{Gao et~al.(2021)Gao, Biderman, Black, Golding, Hoppe, Foster, Phang, He, Thite, Nabeshima, Presser, and Leahy}]{Gao2021Pile}
Leo Gao, Stella Biderman, Sid Black, Laurence Golding, Travis Hoppe, Charles Foster, Jason Phang, Horace He, Anish Thite, Noa Nabeshima, Shawn Presser, and Connor Leahy. 2021.
\newblock \href {https://arxiv.org/abs/2101.00027} {The pile: An 800gb dataset of diverse text for language modeling}.
\newblock \emph{CoRR}, abs/2101.00027.

\bibitem[{Gao et~al.(2024)Gao, la~Tour, Tillman, Goh, Troll, Radford, Sutskever, Leike, and Wu}]{Gao:2024}
Leo Gao, Tom~Dupr{\'{e}} la~Tour, Henk Tillman, Gabriel Goh, Rajan Troll, Alec Radford, Ilya Sutskever, Jan Leike, and Jeffrey Wu. 2024.
\newblock \href {https://doi.org/10.48550/ARXIV.2406.04093} {Scaling and evaluating sparse autoencoders}.
\newblock \emph{CoRR}, abs/2406.04093.

\bibitem[{Geiger et~al.(2024{\natexlab{a}})Geiger, Ibeling, Zur, Chaudhary, Chauhan, Huang, Arora, Wu, Goodman, Potts, and Icard}]{geiger2024causalabstractiontheoreticalfoundation}
Atticus Geiger, Duligur Ibeling, Amir Zur, Maheep Chaudhary, Sonakshi Chauhan, Jing Huang, Aryaman Arora, Zhengxuan Wu, Noah Goodman, Christopher Potts, and Thomas Icard. 2024{\natexlab{a}}.
\newblock \href {https://arxiv.org/abs/2301.04709} {Causal abstraction: A theoretical foundation for mechanistic interpretability}.
\newblock \emph{Preprint}, arXiv:2301.04709.

\bibitem[{Geiger et~al.(2021)Geiger, Lu, Icard, and Potts}]{geiger2021}
Atticus Geiger, Hanson Lu, Thomas Icard, and Christopher Potts. 2021.
\newblock \href {https://proceedings.neurips.cc/paper/2021/hash/4f5c422f4d49a5a807eda27434231040-Abstract.html} {Causal abstractions of neural networks}.
\newblock In \emph{Advances in Neural Information Processing Systems 34: Annual Conference on Neural Information Processing Systems 2021, NeurIPS 2021, December 6-14, 2021, virtual}, pages 9574--9586.

\bibitem[{Geiger et~al.(2020)Geiger, Richardson, and Potts}]{geiger-etal:2020:blackbox}
Atticus Geiger, Kyle Richardson, and Christopher Potts. 2020.
\newblock \href {https://doi.org/10.18653/v1/2020.blackboxnlp-1.16} {Neural natural language inference models partially embed theories of lexical entailment and negation}.
\newblock In \emph{Proceedings of the Third BlackboxNLP Workshop on Analyzing and Interpreting Neural Networks for NLP}, pages 163--173, Online. Association for Computational Linguistics.

\bibitem[{Geiger et~al.(2024{\natexlab{b}})Geiger, Wu, Potts, Icard, and Goodman}]{DAS}
Atticus Geiger, Zhengxuan Wu, Christopher Potts, Thomas Icard, and Noah~D. Goodman. 2024{\natexlab{b}}.
\newblock \href {https://proceedings.mlr.press/v236/geiger24a.html} {Finding alignments between interpretable causal variables and distributed neural representations}.
\newblock In \emph{Causal Learning and Reasoning, 1-3 April 2024, Los Angeles, California, {USA}}, volume 236 of \emph{Proceedings of Machine Learning Research}, pages 160--187. {PMLR}.

\bibitem[{Geva et~al.(2023)Geva, Bastings, Filippova, and Globerson}]{geva2023dissecting}
Mor Geva, Jasmijn Bastings, Katja Filippova, and Amir Globerson. 2023.
\newblock \href {https://arxiv.org/abs/2304.14767} {Dissecting recall of factual associations in auto-regressive language models}.
\newblock \emph{Preprint}, arXiv:2304.14767.

\bibitem[{Geva et~al.(2021)Geva, Schuster, Berant, and Levy}]{geva-2021-transformer}
Mor Geva, Roei Schuster, Jonathan Berant, and Omer Levy. 2021.
\newblock \href {https://aclanthology.org/2021.emnlp-main.446} {Transformer feed-forward layers are key-value memories}.
\newblock In \emph{emnlp}.

\bibitem[{Gurnee et~al.(2023)Gurnee, Nanda, Pauly, Harvey, Troitskii, and Bertsimas}]{polysemantic1}
Wes Gurnee, Neel Nanda, Matthew Pauly, Katherine Harvey, Dmitrii Troitskii, and Dimitris Bertsimas. 2023.
\newblock \href {https://arxiv.org/abs/2305.01610} {Finding neurons in a haystack: Case studies with sparse probing}.
\newblock \emph{Preprint}, arXiv:2305.01610.

\bibitem[{Hernandez et~al.(2023)Hernandez, Li, and Andreas}]{hernandez2023measuring}
Evan Hernandez, Belinda~Z Li, and Jacob Andreas. 2023.
\newblock Measuring and manipulating knowledge representations in language models.
\newblock \emph{arXiv preprint arXiv:2304.00740}.

\bibitem[{Hernandez et~al.(2022)Hernandez, Schwettmann, Bau, Bagashvili, Torralba, and Andreas}]{Hernandez:2022NaturalLanguage}
Evan Hernandez, Sarah Schwettmann, David Bau, Teona Bagashvili, Antonio Torralba, and Jacob Andreas. 2022.
\newblock \href {https://openreview.net/forum?id=NudBMY-tzDr} {Natural language descriptions of deep visual features}.
\newblock In \emph{The Tenth International Conference on Learning Representations, {ICLR} 2022, Virtual Event, April 25-29, 2022}. OpenReview.net.

\bibitem[{Huang et~al.(2023)Huang, Geiger, D'Oosterlinck, Wu, and Potts}]{Huang:2023}
Jing Huang, Atticus Geiger, Karel D'Oosterlinck, Zhengxuan Wu, and Christopher Potts. 2023.
\newblock \href {https://doi.org/10.18653/V1/2023.BLACKBOXNLP-1.24} {Rigorously assessing natural language explanations of neurons}.
\newblock In \emph{Proceedings of the 6th BlackboxNLP Workshop: Analyzing and Interpreting Neural Networks for NLP, BlackboxNLP@EMNLP 2023, Singapore, December 7, 2023}, pages 317--331. Association for Computational Linguistics.

\bibitem[{Huang et~al.(2024)Huang, Wu, Potts, Geva, and Geiger}]{ravel}
Jing Huang, Zhengxuan Wu, Christopher Potts, Mor Geva, and Atticus Geiger. 2024.
\newblock \href {https://arxiv.org/abs/2402.17700} {Ravel: Evaluating interpretability methods on disentangling language model representations}.
\newblock \emph{Preprint}, arXiv:2402.17700.

\bibitem[{Lieberum et~al.(2024)Lieberum, Rajamanoharan, Conmy, Smith, Sonnerat, Varma, Kramár, Dragan, Shah, and Nanda}]{lieberum2024gemmascopeopensparse}
Tom Lieberum, Senthooran Rajamanoharan, Arthur Conmy, Lewis Smith, Nicolas Sonnerat, Vikrant Varma, János Kramár, Anca Dragan, Rohin Shah, and Neel Nanda. 2024.
\newblock \href {https://arxiv.org/abs/2408.05147} {Gemma scope: Open sparse autoencoders everywhere all at once on gemma 2}.
\newblock \emph{Preprint}, arXiv:2408.05147.

\bibitem[{Makelov et~al.(2024)Makelov, Lange, and Nanda}]{makelov2024}
Aleksandar Makelov, George Lange, and Neel Nanda. 2024.
\newblock \href {https://arxiv.org/abs/2405.08366} {Towards principled evaluations of sparse autoencoders for interpretability and control}.
\newblock \emph{Preprint}, arXiv:2405.08366.

\bibitem[{Marks et~al.(2024)Marks, Rager, Michaud, Belinkov, Bau, and Mueller}]{marks2024}
Samuel Marks, Can Rager, Eric~J. Michaud, Yonatan Belinkov, David Bau, and Aaron Mueller. 2024.
\newblock \href {https://doi.org/10.48550/ARXIV.2403.19647} {Sparse feature circuits: Discovering and editing interpretable causal graphs in language models}.
\newblock \emph{CoRR}, abs/2403.19647.

\bibitem[{McClelland et~al.(1986{\natexlab{a}})McClelland, Rumelhart, and Group}]{polysemantic6}
James~L. McClelland, David~E. Rumelhart, and PDP~Research Group. 1986{\natexlab{a}}.
\newblock \href {https://doi.org/10.7551/mitpress/5237.001.0001} {\emph{Parallel Distributed Processing, Volume 2: Explorations in the Microstructure of Cognition: Psychological and Biological Models}}.
\newblock The MIT Press.

\bibitem[{McClelland et~al.(1986{\natexlab{b}})McClelland, Rumelhart, and Group}]{polysemantic5}
James~L. McClelland, David~E. Rumelhart, and PDP~Research Group. 1986{\natexlab{b}}.
\newblock \href {https://doi.org/10.7551/mitpress/5237.001.0001} {\emph{Parallel Distributed Processing, Volume 2: Explorations in the Microstructure of Cognition: Psychological and Biological Models}}.
\newblock The MIT Press.

\bibitem[{Meng et~al.(2022)Meng, Bau, Andonian, and Belinkov}]{Meng:2022}
Kevin Meng, David Bau, Alex Andonian, and Yonatan Belinkov. 2022.
\newblock \href {http://papers.nips.cc/paper\_files/paper/2022/hash/6f1d43d5a82a37e89b0665b33bf3a182-Abstract-Conference.html} {Locating and editing factual associations in {GPT}}.
\newblock In \emph{Advances in Neural Information Processing Systems 35: Annual Conference on Neural Information Processing Systems 2022, NeurIPS 2022, New Orleans, LA, USA, November 28 - December 9, 2022}.

\bibitem[{Meng et~al.(2023)Meng, Sharma, Andonian, Belinkov, and Bau}]{Meng:2023}
Kevin Meng, Arnab~Sen Sharma, Alex~J. Andonian, Yonatan Belinkov, and David Bau. 2023.
\newblock \href {https://openreview.net/forum?id=MkbcAHIYgyS} {Mass-editing memory in a transformer}.
\newblock In \emph{The Eleventh International Conference on Learning Representations, {ICLR} 2023, Kigali, Rwanda, May 1-5, 2023}. OpenReview.net.

\bibitem[{Mueller et~al.(2024)Mueller, Brinkmann, Li, Marks, Pal, Prakash, Rager, Sankaranarayanan, Sharma, Sun, Todd, Bau, and Belinkov}]{mueller2024questrightmediatorhistory}
Aaron Mueller, Jannik Brinkmann, Millicent Li, Samuel Marks, Koyena Pal, Nikhil Prakash, Can Rager, Aruna Sankaranarayanan, Arnab~Sen Sharma, Jiuding Sun, Eric Todd, David Bau, and Yonatan Belinkov. 2024.
\newblock \href {https://arxiv.org/abs/2408.01416} {The quest for the right mediator: A history, survey, and theoretical grounding of causal interpretability}.
\newblock \emph{Preprint}, arXiv:2408.01416.

\bibitem[{Olah et~al.(2020)Olah, Cammarata, Schubert, Goh, Petrov, and Carter}]{polysemantic4}
Chris Olah, Nick Cammarata, Ludwig Schubert, Gabriel Goh, Michael Petrov, and Shan Carter. 2020.
\newblock \href {https://doi.org/10.23915/distill.00024.001} {Zoom in: An introduction to circuits}.
\newblock \emph{Distill}.
\newblock Https://distill.pub/2020/circuits/zoom-in.

\bibitem[{Paszke et~al.(2019)Paszke, Gross, Massa, Lerer, Bradbury, Chanan, Killeen, Lin, Gimelshein, Antiga, Desmaison, Kopf, Yang, DeVito, Raison, Tejani, Chilamkurthy, Steiner, Fang, Bai, and Chintala}]{pytorch}
Adam Paszke, Sam Gross, Francisco Massa, Adam Lerer, James Bradbury, Gregory Chanan, Trevor Killeen, Zeming Lin, Natalia Gimelshein, Luca Antiga, Alban Desmaison, Andreas Kopf, Edward Yang, Zachary DeVito, Martin Raison, Alykhan Tejani, Sasank Chilamkurthy, Benoit Steiner, Lu~Fang, Junjie Bai, and Soumith Chintala. 2019.
\newblock \href {http://papers.neurips.cc/paper/9015-pytorch-an-imperative-style-high-performance-deep-learning-library.pdf} {Pytorch: An imperative style, high-performance deep learning library}.
\newblock In \emph{Advances in Neural Information Processing Systems 32}, pages 8024--8035. Curran Associates, Inc.

\bibitem[{Radford et~al.(2019)Radford, Wu, Child, Luan, Amodei, and Sutskever}]{radford2019language}
Alec Radford, Jeff Wu, Rewon Child, David Luan, Dario Amodei, and Ilya Sutskever. 2019.
\newblock Language models are unsupervised multitask learners.

\bibitem[{Schwettmann et~al.(2023)Schwettmann, Shaham, Materzynska, Chowdhury, Li, Andreas, Bau, and Torralba}]{Schwettmann2023FIND}
Sarah Schwettmann, Tamar~Rott Shaham, Joanna Materzynska, Neil Chowdhury, Shuang Li, Jacob Andreas, David Bau, and Antonio Torralba. 2023.
\newblock \href {http://papers.nips.cc/paper\_files/paper/2023/hash/ef0164c1112f56246224af540857348f-Abstract-Datasets\_and\_Benchmarks.html} {{FIND:} {A} function description benchmark for evaluating interpretability methods}.
\newblock In \emph{Advances in Neural Information Processing Systems 36: Annual Conference on Neural Information Processing Systems 2023, NeurIPS 2023, New Orleans, LA, USA, December 10 - 16, 2023}.

\bibitem[{Shaham et~al.(2024)Shaham, Schwettmann, Wang, Rajaram, Hernandez, Andreas, and Torralba}]{Shaham2024}
Tamar~Rott Shaham, Sarah Schwettmann, Franklin Wang, Achyuta Rajaram, Evan Hernandez, Jacob Andreas, and Antonio Torralba. 2024.
\newblock \href {https://doi.org/10.48550/ARXIV.2404.14394} {A multimodal automated interpretability agent}.
\newblock \emph{CoRR}, abs/2404.14394.

\bibitem[{Smolensky(1988)}]{polysemantic7}
Paul Smolensky. 1988.
\newblock \href {https://doi.org/10.1017/S0140525X00052432} {On the proper treatment of connectionism}.
\newblock \emph{Behavioral and Brain Sciences}, 11(1):1–23.

\bibitem[{Templeton et~al.(2024)Templeton, Conerly, Marcus, Lindsey, Bricken, Chen, Pearce, Citro, Ameisen, Jones, Cunningham, Turner, McDougall, MacDiarmid, Freeman, Sumers, Rees, Batson, Jermyn, Carter, Olah, and Henighan}]{templeton2024scaling}
Adly Templeton, Tom Conerly, Jonathan Marcus, Jack Lindsey, Trenton Bricken, Brian Chen, Adam Pearce, Craig Citro, Emmanuel Ameisen, Andy Jones, Hoagy Cunningham, Nicholas~L Turner, Callum McDougall, Monte MacDiarmid, C.~Daniel Freeman, Theodore~R. Sumers, Edward Rees, Joshua Batson, Adam Jermyn, Shan Carter, Chris Olah, and Tom Henighan. 2024.
\newblock \href {https://transformer-circuits.pub/2024/scaling-monosemanticity/index.html} {Scaling monosemanticity: Extracting interpretable features from claude 3 sonnet}.
\newblock \emph{Transformer Circuits Thread}.

\bibitem[{Vig et~al.(2020)Vig, Gehrmann, Belinkov, Qian, Nevo, Singer, and Shieber}]{vig2020}
Jesse Vig, Sebastian Gehrmann, Yonatan Belinkov, Sharon Qian, Daniel Nevo, Yaron Singer, and Stuart~M. Shieber. 2020.
\newblock \href {https://proceedings.neurips.cc/paper/2020/hash/92650b2e92217715fe312e6fa7b90d82-Abstract.html} {Investigating gender bias in language models using causal mediation analysis}.
\newblock In \emph{Advances in Neural Information Processing Systems 33: Annual Conference on Neural Information Processing Systems 2020, NeurIPS 2020, December 6-12, 2020, virtual}.

\end{thebibliography}

\newpage

\appendix

\section{Evaluation Details}\label{app:eval}

To enhance, the prediction capability of GPT-$2$ using in-context learning, we use 5-shot prompt for both the attributes. 
Specifically, for country attribute, we prepare a template as:
\emph{``Toronto is a city in the country of Canada. Beijing is a city in the country of China. Miami is a city in the country of the United States. Santiago is a city in the country of Chile. London is a city in the country of England. \texttt{<city>} is a city in the country of''}.

Similarly, to support the prediction of continent, we also prepare a similar template for the model as:
\emph{``Toronto is a city in the continent of North America. Beijing is a city in the continent of Asia. Miami is a city in the continent of North America. Santiago is a city in the continent of South America. London is a city in the continent of Europe. \texttt{<city>} is a city in the continent of''}.
The \texttt{<city>} is replaced with the city name in the dataset to make several samples to make the data for both the country and continent attributes.

Eventually, we prepare the final dataset consisting of base and source sentences, with their corresponding labels to evaluate different techniques. 
In each example, either the `country' is targeted for intervention or the `continent' is. When a prompt is for targeted attribute, the intervention should change the output to match the source city. When the prompt is for the other attribute, the intervention should not change the output.

\section{Hyperparameters and Compute}\label{app:hyper}
We used these parameters for DBM and DBM+DAS training. Batch size of 16. Temperature is annealed linearly from $10$ to  $0.1$. Training was for $20$ epochs. Learning rate is $0.001$. 

A masking experiment takes 1 hour approx to run. Three layers had 4 experiments with a run for for each intervention so 4*2 experiments.
Layer 1 had a total of 6 experiments with two interventions each.
Total time: 1x3x4x2 + 1x6x2 = 36 hours on a 24GB Nvidia RTX A5000

\section{Full Reconstruction Evaluation}
See Tables~\ref{table:reconloss} and~\ref{table:reconperformance} for the reconstruction evaluations done across all the layers.

\begin{table*}[ht]
 \centering
 \resizebox{\textwidth}{!}{
 \begin{tabular}{|c | c | c | c | c | c | c| c| c|}
 \hline
 Layers & Bloom SAE & Bloom SAE & OpenAI SAE & 
 OpenAI SAE & Apollo SAE & Apollo SAE & Apollo SAE & Apollo SAE\\
 & Country & Continent & Country & Continent & Country & Continent &  e2e+ds Country & e2e+ds Continent\\
 \hline
 Layer 0 & 400.87 & 413.03 & 102.91 & 104.2 &  - & - & - & -  \\
 \hline
 Layer 1 & 551.28 & 516.5 & 151.83 & 158.03 & 2245.57 & 2307.15  & 2129.71 & 2123.09\\
 \hline
 Layer 2 & 698.25 & 681.64 & 217.13 & 219.78 & - &  -  & - & -\\
 \hline
 Layer 3 & 876.36 & 814.99 & 330.43 & 336.34 & - & -  & - & - \\
 \hline
 Layer 4 & 890.41 & 869.71 & 449.33 & 458.82 & - & -  & - & - \\
 \hline
 Layer 5 & 936.77 & 1044.33 & 643.82 & 651.67 & 2383.14 & 2576.08  & 2353.61 & 2318.49\\
 \hline
 Layer 6 & 1178.01 & 1531.46 & 839.68 & 837.81 & - & -  & - & - \\
 \hline
 Layer 7 & 4640.78 & 7757.06 & 1218.99 & 1211.81 & - & - & - & - \\
 \hline
 Layer 8 & 19556.78 & 26810.38 & 1727.77 & 1723.93 & - & - & - & -\\
 \hline
 Layer 9 & 27877.84 & 36537.93 & 2304.84 & 2311.26 & 5276.6 & 6038.87  & 2569.59 & 2665.5\\
 \hline
 Layer 10 & 532812.74 & 571233.39 & 3296.77 & 3467.73 & - & - & - & -\\
 \hline
 Layer 11 & 846887.04 & 859555.3 & 4833.99 & 4893.55 & - & - & - & -\\
 \hline
 \end{tabular}}
 \caption{The table above denotes the reconstruction loss for country and continent dataset separately for each SAE.}
 \label{table:reconloss}
 \end{table*}

\begin{table*}[ht]
\centering
\resizebox{\textwidth}{!}{
\begin{tabular}{|c | c | c | c | c | c | c|c|c|}
\hline
Layers & Bloom SAE & Bloom SAE & OpenAI SAE & 
 OpenAI SAE & Apollo SAE & Apollo SAE & Apollo SAE & Apollo SAE \\
 & Country & Continent & Country & Continent & Country & Continent & e2e+ds Country & e2e+ds Continent \\
\hline
 Layer 0 & 0.9375 & 0.890625 & 0.9642857142857143 & 0.9609375 & 
 -&- & -& -\\
 \hline
 Layer 1 & 0.5625 & 0.46875 & 0.9464285714285714 & 0.9453125 & 
 0.6696428571428571 & 0.3515625  &0.008928571428571428 & 0.0\\
 \hline
 Layer 2 & 0.5267857142857143 & 0.5390625 & 0.9553571428571429 & 0.9140625 & -
  & - & - & -\\
 \hline
 Layer 3 & 0.7142857142857143 & 0.78125 & 0.9196428571428571 & 0.890625 & - 
  & - & - & -\\
 \hline
 Layer 4 & 0.7946428571428571 & 0.8984375 & 0.9196428571428571 & 0.875 & 
 -& - & - & -\\
 \hline
 Layer 5 & 0.7678571428571429 & 0.859375 & 0.875 & 0.8984375 
  & 0.8392857142857143 & 0.7578125 &0.8303571428571429 & 0.609375 \\
 \hline
 Layer 6 & 0.7946428571428571 & 0.78125 & 0.8125 & 0.7421875 & - & - & - & -
 \\
 \hline
 Layer 7 & 0.875 & 0.765625 & 0.7857142857142857 & 0.703125 & - & - & - & -\\
 \hline
 Layer 8 & 0.8571428571428571 & 0.7578125 & 0.8571428571428571 & 0.8984375 & - & - & - & - \\
 \hline
 Layer 9 & 0.6696428571428571 & 0.5078125 & 0.9107142857142857 & 0.9609375 & 
 0.9464285714285714 & 0.9453125 &  0.875 & 0.90625\\
 \hline
 Layer 10 & 0.23214285714285715 & 0.03125 & 0.9732142857142857 & 0.9765625 & - & - & - & - \\
 \hline
 Layer 11 & 1.0 & 1.0 & 1.0 & 1.0 & - & - & - & -\\
  \hline
\end{tabular}}
\caption{The table above denotes the accuracy for country and continent dataset after intervention for each SAE}
\label{table:reconperformance}
\end{table*}

\section{Training Graphs}
\begin{figure*}
    \centering
\begin{subfigure}{\textwidth}
    \centering
    \includegraphics[width=0.495\textwidth]{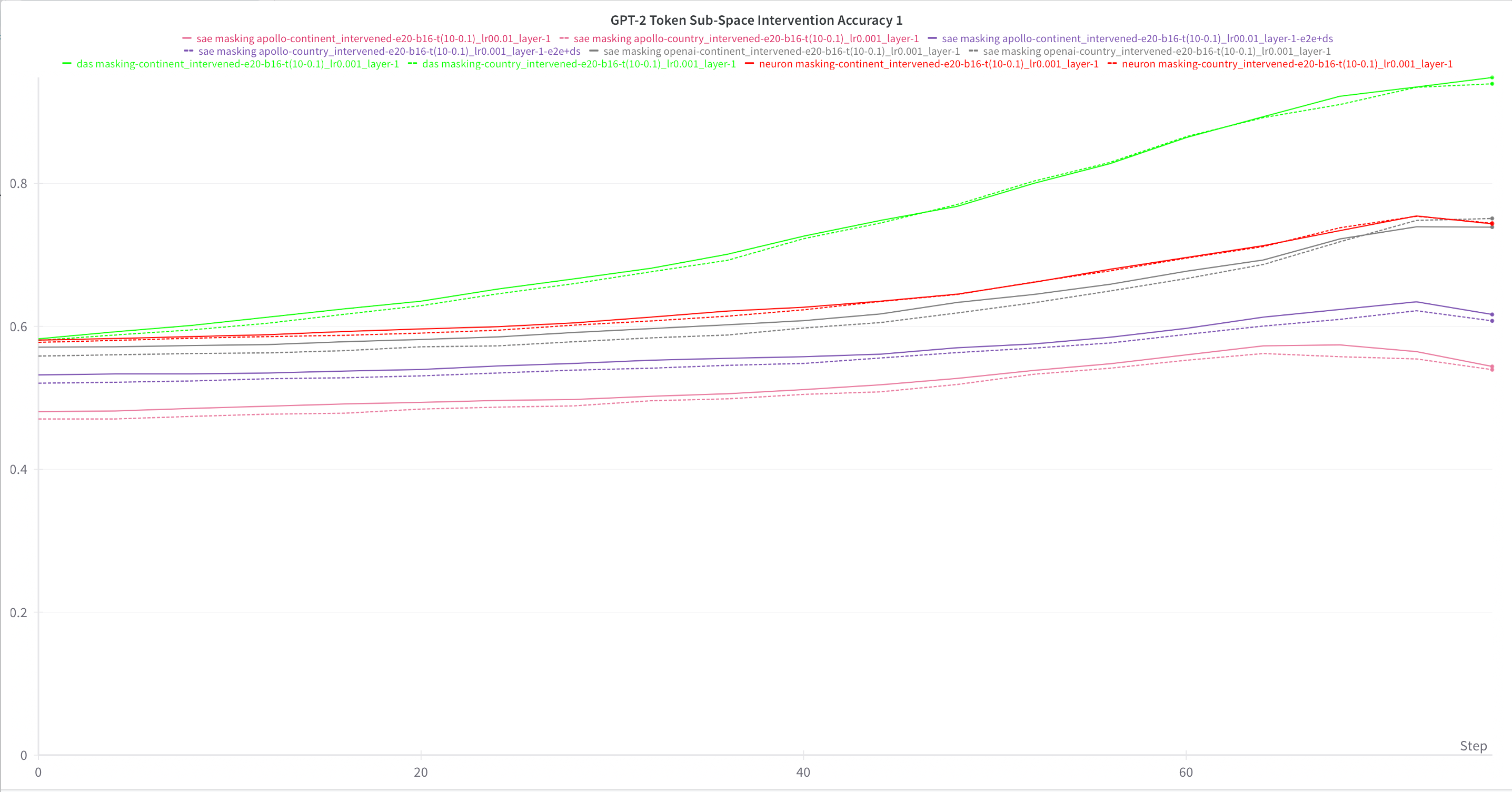}
    \includegraphics[width=0.495\textwidth]{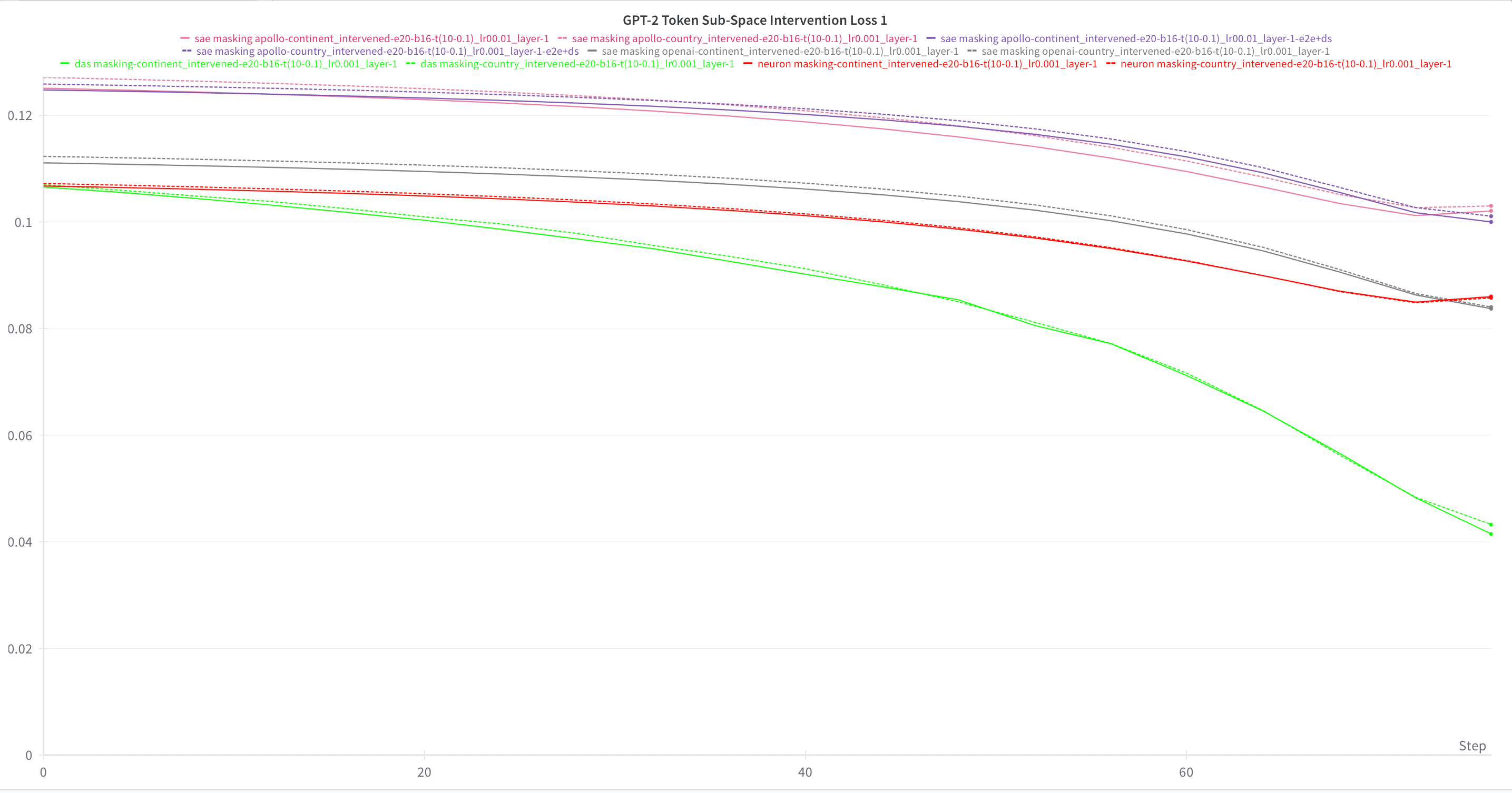}
    \caption{Training graphs for Layer $1$ depict the results for both country and continent interventions. The country-intervened data is represented with dashed lines, while continent-intervened data is shown with bold lines, using the same color scheme as defined in the legend above the graph. The plots illustrate the training accuracy and loss for \textcolor{red}{Neuron Masking}, \textcolor{pink}{SAE Apollo e2e}, \textcolor{violet}{SAE Apollo e2e+ds}, \textcolor{gray}{OpenAI SAE}, and \textcolor{blue}{Bloom SAE} with \textcolor{green}{DAS}. }
    \label{fig:Train_L1}
\end{subfigure}
\end{figure*}

\begin{figure*}[hbtp]
    \centering
\begin{subfigure}{\textwidth}
    \centering
    \includegraphics[width=0.495\textwidth]{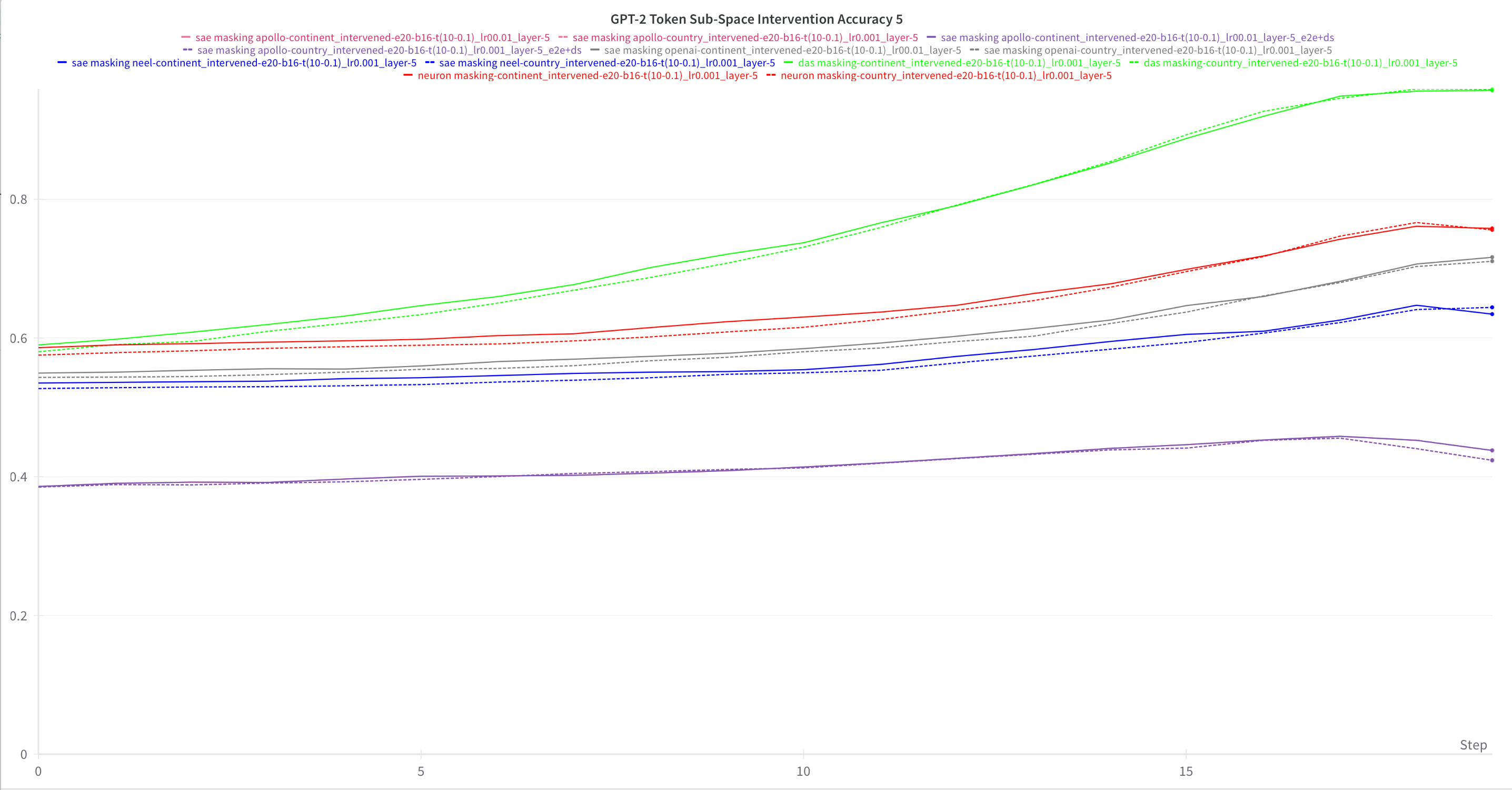}
    \includegraphics[width=0.495\textwidth]{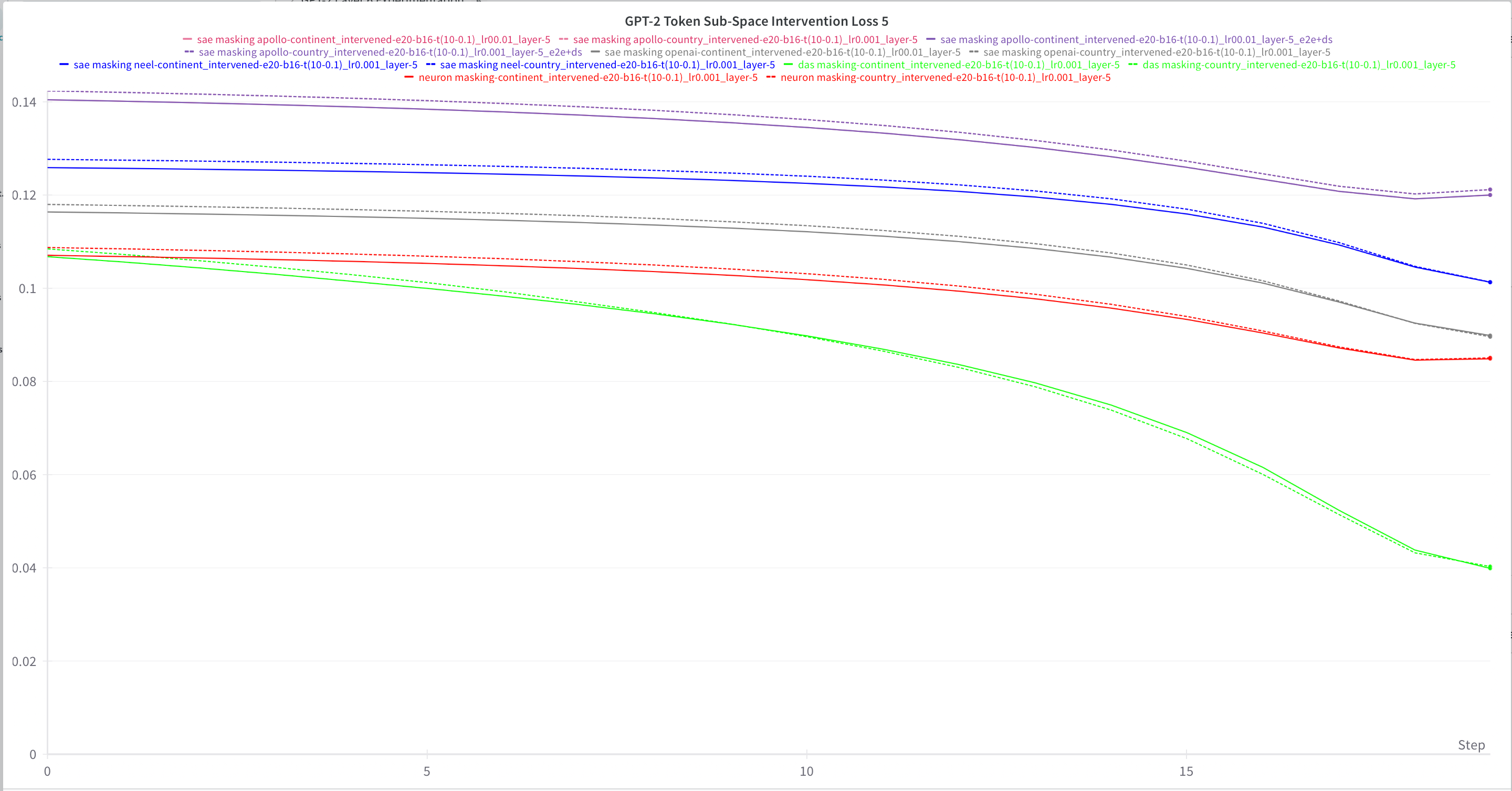}
        \caption{Training graphs for Layer $5$ depict the results for both country and continent interventions. The country-intervened data is represented with dashed lines, while continent-intervened data is shown with bold lines, using the same color scheme as defined in the legend above the graph. The plots illustrate the training accuracy and loss for \textcolor{red}{Neuron Masking}, \textcolor{pink}{SAE Apollo e2e}, \textcolor{violet}{SAE Apollo e2e+ds}, \textcolor{gray}{OpenAI SAE}, and \textcolor{blue}{Bloom SAE} with \textcolor{green}{DAS}. }
    \label{fig:Train_L5}
\end{subfigure}
\end{figure*}

See Figures~\ref{fig:Train_L1} and~\ref{fig:Train_L5} for the training graphs.

\end{document}